\documentclass[conference]{IEEEtran}
\IEEEoverridecommandlockouts
\usepackage{cite}
\usepackage{amsmath,amssymb,amsfonts}
\usepackage{algorithmic}
\usepackage{graphicx}
\usepackage{textcomp}
\usepackage{xcolor}
\usepackage{booktabs}
\def\BibTeX{{\rm B\kern-.05em{\sc i\kern-.025em b}\kern-.08em
    T\kern-.1667em\lower.7ex\hbox{E}\kern-.125emX}}
\begin{document}

\title{Alzheimer’s Dementia Detection through Spontaneous Dialogue with Proactive Robotic Listeners}

\author{\IEEEauthorblockN{Yuanchao Li, Catherine Lai}
\IEEEauthorblockA{\textit{School of Informatics} \\
\textit{University of Edinburgh, UK} \\
y.li-385@sms.ed.ac.uk, c.lai@ed.ac.uk}
\and
\IEEEauthorblockN{Divesh Lala,
Koji Inoue, Tatsuya Kawahara}
\IEEEauthorblockA{\textit{Graduate School of Informatics} \\
\textit{Kyoto University, Japan} \\
\{lala, inoue, kawahara\}@sap.ist.i.kyoto-u.ac.jp}}

\maketitle

\begin{abstract}
As the aging of society continues to accelerate, Alzheimer's Disease (AD) has received more and more attention from not only medical but also other fields, such as computer science, over the past decade. Since speech is considered one of the effective ways to diagnose cognitive decline, AD detection from speech has emerged as a hot topic. Nevertheless, such approaches fail to tackle several key issues: 1) AD is a complex neurocognitive disorder which means it is inappropriate to conduct AD detection using utterance information alone while ignoring dialogue information; 2) Utterances of AD patients contain many disfluencies that affect speech recognition yet are helpful to diagnosis; 3) AD patients tend to speak less, causing dialogue breakdown as the disease progresses. This fact leads to a small number of utterances, which may cause detection bias. Therefore, in this paper, we propose a novel AD detection architecture consisting of two major modules: an ensemble AD detector and a proactive listener. This architecture can be embedded in the dialogue system of conversational robots for healthcare.
\end{abstract}


\section{Introduction}
With the rapid growth of the elderly population, Alzheimer’s Disease (AD) has become a serious problem in today's aging society. AD is a neurodegenerative disease that progressively deteriorates memory, language, and cognitive abilities. Hence, early detection of AD for prevention is the most important to tackle the disease \cite{b1}. Many studies have proved that AD is recognizable from spontaneous speech \cite{b2,b3,b4}, and both audio and transcript information contribute to the detection \cite{b5, b6}. These studies have greatly advanced the early detection of AD.

AD detection consists of two necessary steps: feature extraction and classification/regression. The ADReSS Challenge at INTERSPEECH 2020 \cite{b7} has presented a comparison study using several baseline feature sets, which include \textit{emobase} \cite{b8}, \textit{ComParE} \cite{b9}, \textit{eGeMAPS} \cite{b10}, \textit{MRCG functionals} \cite{b11}, and \textit{Minimal} \cite{b12}. These feature sets include acoustic features such as energy, Mel-Frequency Cepstral Coefficients (MFCC), fundamental frequency (F0), and so on, as well as their statistical functionals. Besides, directly learning the mapping from raw speech signals using neural networks has emerged as a trend in current work \cite{b13}. Traditional machine learning approaches such as linear discriminant analysis, decision trees, nearest neighbor, random forests, and support vector machines are adopted for classification and regression using extracted features in the ADReSS Challenge. In addition, deep learning-based approaches are also being investigated for the same tasks and show great performance \cite{b14, b15}.

Despite the progress, AD detection remains challenging due to several key issues. First of all, AD is a complex neurocognitive disorder that needs a professional medical diagnosis. Detection using computers with only utterance-level information is inappropriate and causes inaccurate results \cite{b3}. Thus, more information from dialogue aspects (e.g., turn-taking time) should be considered. Second, AD patients are usually not able to speak as fluently and clearly as non-AD people. Their utterances contain many disfluencies, such as fillers, false starts, repetitions, and so on \cite{b15, b16}. These disfluencies affect Automatic Speech Recognition (ASR) performance, and as a consequence, the error-prone ASR transcripts are not as reliable as manual transcripts. Third, AD patients tend to speak less as the disease progresses, which leads to a small number of utterances \cite{b17}. In that case, data scarcity can cause a serious detection bias problem.

Nowadays, robots with a dialogue function have been replacing human labor in several scenarios related to reception, presentation, elderly care, and the like \cite{b18}. Therefore, we propose a novel AD detection architecture that is expected to solve the above-mentioned problems. This proposed architecture can be embedded in the dialogue system of conversational robots, with the aim of breaking disciplinary boundaries to help AD patients.

\section{Related Work}

\subsection{AD Detection from Speech}
To automate AD detection, researchers have used various acoustic and lexical features. Traditionally, a number of acoustic-related low-level descriptors built upon prior knowledge have been handcrafted to represent AD. \cite{b4} evaluated several handcrafted feature sets designed for different computational paralinguistics tasks and proposed a novel Active Data Representation method using acoustic features of all speech segments with a single fixed-dimension feature vector. Inspired by successful end-to-end approaches in speech and emotion recognition \cite{b19, b20}, some recent works extract features directly from log-mel spectrograms instead of using handcrafted ones, resulting in better performance \cite{b21, b22}.

Unlike from acoustic analysis, language research employs high-level lexical features for AD detection. For example, Lu's L2 Syntactic Complexity Analyzer computed 23 features that measure the syntactic complexity of the text, which include lengths of production units, the ratio of clauses to sentences, subordination, coordination, and particular structures \cite{b23}. Thanks to the rapid development of Natural Language Processing (NLP), current research is adopting pre-trained models such as GloVe \cite{b24}, BERT \cite{b25}, and RoBERTa \cite{b26} to extract lexical features \cite{b13, b15, b27}, and such methods usually outperform handcrafted feature engineering \cite{b28}.

Alongside acoustic and lexical features, some work also takes into consideration interactivity and disfluency features. Based on the finding that therapist-patient dialogues can be regarded as Markov chains \cite{b29}, \cite{b30} presented descriptive statistics on dialogues, such as dialogue duration, turn duration, number of words, and words per minute. Disfluencies, a natural pattern in spontaneous dialogues, are necessary for detecting cognitive diseases \cite{b31}, which leads to AD detection using self-repairs, editing terms, short pauses, and long pauses \cite{b14}.

In this paper, we follow prior work and propose a novel detection approach that uses two-stage ensemble learning to integrate multiple classifiers for the final AD detection results.

\subsection{Conversational Robots for Healthcare}
Conversational robots have been used in various situations nowadays, and they have proven the importance of social Human-Robot Interaction (HRI) as a means of providing effective healthcare. Prior research has demonstrated that the elderly are eager to talk to robots because of the lack of social ties. Even simple social conventions like daily greetings from the robot can comfort the elderly with the assurance that the robot will always be there to respond \cite{b32}. This finding provides a basis to use conversational robots for AD detection. Recent work has investigated experiment design and strategies for AD detection using conversational robots. \cite{b33} indicated that even though the conversational robot presents a non-pharmacological treatment approach, its daily use can have a therapeutic effect on the behavioral and psychological symptoms of dementia in older adults. \cite{b34} explored how conversational robots can be used to support individuals with AD, and their results show that the robot was generally well-liked by AD patients and that it could capture their interest in dialogue. They also highlighted the robot's potential as a monitoring tool by analyzing how non-acoustic aspects of language change across participants with different AD degrees. \cite{b35} developed a prototype of a listener agent, and collected conversations between the agent and people with dementia for assessment using a conversation log system.

However, the above-mentioned research ignores the fact that AD patients tend to speak less as the disease progresses. This dialogue breakdown impedes the automatic AD detection from being applied in real life. Hence, there is a need for the system to maintain the conversation until an overall diagnosis is conducted. In this paper, we propose to use a proactive listener to resolve this issue.

\section{Proposed Approach}
\subsection{Architecture Overview}
The overview of the proposed architecture is shown in Fig. 1, and the processing steps are as follows. First, the spontaneous dialogue between the human and the robot takes place regardless of who starts it. The human’s speech signals are perceived by the robot’s microphones and then processed by the front end. It should be noted that to process the speech signals, there are two paths, one of which directly extracts acoustic features and models the feature sequence, and the other converts speech into ASR transcripts. We choose to use open-source tools to perform this step because they are convenient to implement in programmable robots.

Next, \textit{proactive listener}, one of the major modules in the architecture, recognizes the type of user speech with a focus word extractor and a dialogue act tagger. This proactive listener is a variant of a previously proposed attentive listener \cite{b36}, mainly by removing backchanneling and flexible turn-taking functions and adding a proactive initiator \cite{b37}. We define user speech into three types: question, statement, and silence. Each type has its corresponding response: answer, question, partial repeat, follow-up question, and topic introduction. The proactive responses aim to maintain the conversation to acquire more speech samples from the human user for a comprehensive diagnosis.

Meanwhile, \textit{ensemble AD detector}, the other major module in the architecture, detects the AD degree using a two-stage ensemble learning approach. Inside this module, there are four classifiers conducting AD detection on four different modalities respectively: audio, language, disfluency, and interactivity. Then an ensemble learning model applies averaging and a majority vote in two stages on the four probability distributions to obtain the final decision. The output has four degrees: mild AD, moderate AD, severe AD, and non-AD, which will finally be recorded in medical logs for medical professionals to follow up on the disease conditions in long term.

\begin{figure*}[htbp]
\centering
\includegraphics[width=0.88\textwidth]{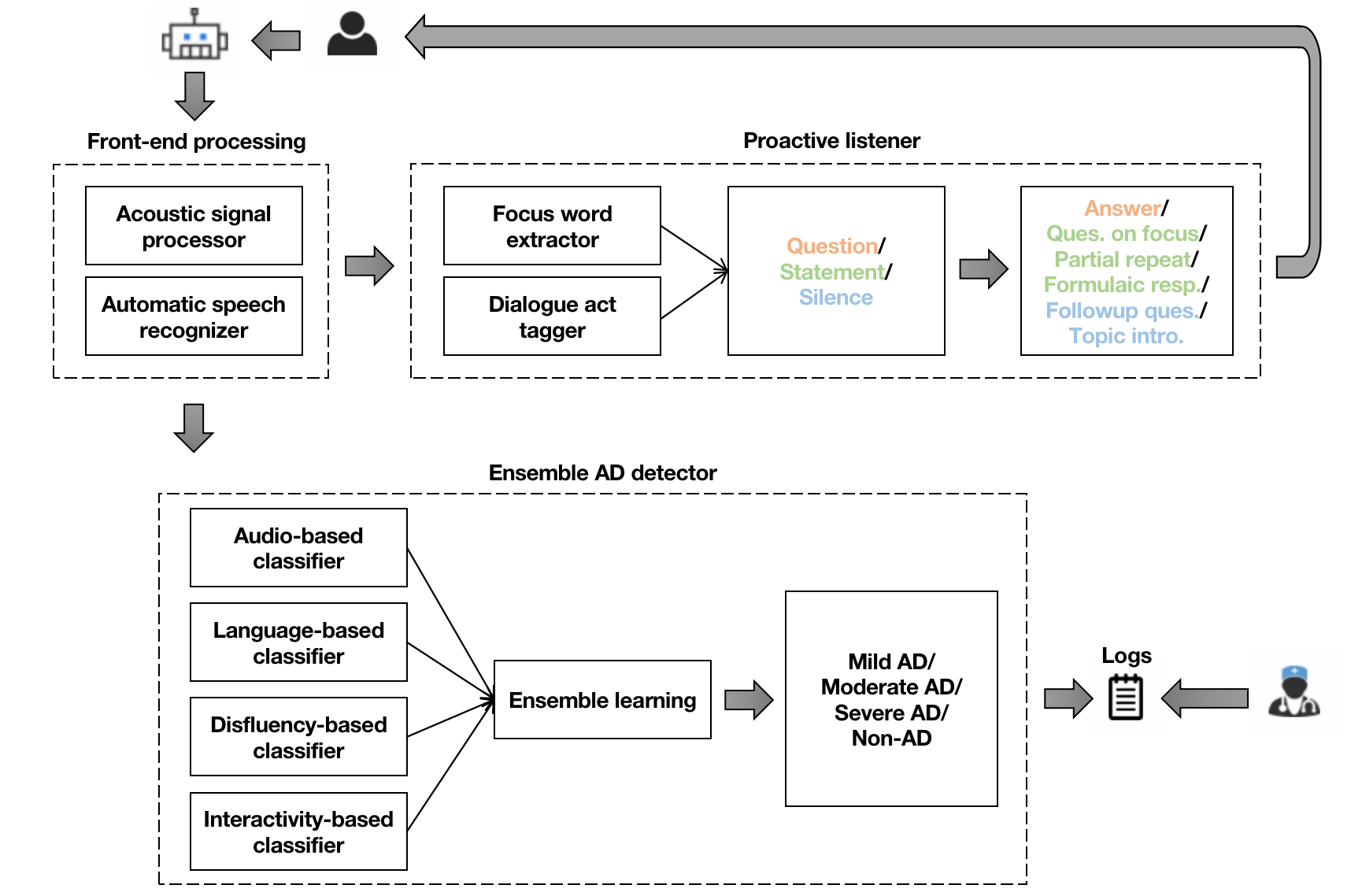}
\caption{The proposed AD detection architecture.}
\end{figure*}

\subsection{Detail Description}
In this section, we give a detailed description of the two major modules.

\subsubsection{Proactive Listener}
Our goal of implementing the proactive listener is to maintain spontaneous dialogues between the robot and the human, considering that the dialogues are easy to break down since AD patients tend to speak less as the disease progresses. The focus word extractor is one of the NLP functionalities that identifies the focus of an utterance based on a conditional random field classifier that uses part-of-speech tags and a phrase-level dependency tree \cite{b38}. The dialogue act tagger classifies user utterances into question, statement or silence, and works together with the focus word extractor.

For ``question'', the response will be generated based on adjacency pairs from a handcrafted question-answer database. Compared to statistical methods (e.g., neural dialogue models) for generating responses, carefully crafted rule-based system is better in this situation for control, since dialogues with AD patients are vulnerable. For ``statement'', the response can be a question or a partial repeat according to a decision tree scheme. We follow a previous approach using an n-gram language model to compute the joint probability of the focus noun being associated with each wh-question word (i.e., who, what, when, where) \cite{b36}. If the maximum joint probability of this noun and a question word is over a pre-defined threshold, then a question on the focus word is generated. Otherwise, a partial repeat is generated. If no focus noun, then a formulaic response is generated for rapport. For ``silence'', if it lasts five seconds, then a follow-up question related to the most recent topic will be responded. If the silence continues lasting another five seconds, then a topic introduction will be started. A dialogue example is shown in Table. 1.

\begin{table}
\centering
\caption{A dialogue example consisting of different utterance-response types. H: Human, R: Robot}
\begin{tabular}{lll}
\toprule
\textbf{Utterance-response type} & \textbf{Example} \\ \midrule
Question-answer & H: How is the weather? \\
 & R: It's raining outside. \\ \midrule
Statement-ques. on focus & H: OK, I'll watch a \underline{movie} then. \\
 & R: Which movie? \\ \midrule
Statement-partial repeat & H: \underline{Avengers}, the newest one. \\
 & R: Avengers? \\ \midrule
Silence-followup ques. & H: [5s silence] \\
 & R: What's your favorite movie? \\ \midrule
Silence-topic intro. & H: [5s silence] \\
 & R: Do you like music? \\ \midrule
Statement-formulaic resp. & H: Yes, I like. \\
 & R: That's good. \\ \bottomrule
\end{tabular}
\end{table}

\subsubsection{Ensemble AD Detector}
The ensemble AD detector conducts detection on four aspects: audio, language, disfluency, and interactivity.

I. \textit{Audio-based classifier.} Following a prior HRI work \cite{b39}, we first use the OpenSMILE toolbox \cite{b40} to extract ComParE \cite{b9} acoustic feature set as a baseline. We will also directly use raw speech in an end-to-end manner for comparison. In addition, we will also use a pre-trained acoustic model such as wav2vec \cite{b41} or HuBERT \cite{b42} to explore its efficiency in real-time application. We present the sequence of the extracted acoustic feature vectors as [$a_1$, ..., $a_M$], where $M$ is the sequence length.

II. \textit{Language-based classifier.} We use word embedding to convert each word from the text as an embedding vector. We will also use a pre-trained language model such as GloVe \cite{b24}, BERT \cite{b25} or RoBERTa \cite{b26} for the same reason above. We present the sequence of the extracted lexical feature vectors as [$l_1$, ..., $l_N$], where $N$ is the sequence length.

III. \textit{Disfluency-based classifier.} We categorize disfluency into four categories: restart, repetition, correction, and filler. The traditional way to detect AD from disfluency is first predicting disfluency from feature inputs using SVMs or neural networks, and then detecting AD from disfluency patterns according to a linear regression mapping relationship. However, considering the ASR error in transcripts, which leads to inaccurate disfluency prediction, we directly detect AD from feature inputs. Following \cite{b43}, we use one-hot vectors from the ASR transcripts as the lexical feature input. Following \cite{b44}, we use voice activity, pitch, intensity, and spectral stability as the prosodic feature input. We define the extracted lexical and prosodic features as disfluency features and present their sequence as [$d_1$, ..., $d_P$], where $P$ is the sequence length.

IV. \textit{Interactivity-based classifier.} Unlike other three classifiers that conduct utterance-level detection (one detection result per utterance), interactivity-based classifier conducts dialogue-level detection (one detection result per dialogue). We define six turn-pairs as a dialogue, and extract the following interactional features according to prior knowledge \cite{b45}: turn length (number of words per turn), floor control ratio (time amount during the human speaker speaks to the total speech time of the dialogue), standardized pause rate (ratio of total words to the total pauses), phonation rate (total time spoken to total spoken time including pause), and speaking rate (number of words per minute).

The interactivity-based classifier will be built on linear regression since there are only five feature types. The other three classifiers, as well as the dialogue act tagger, will be built on Gated Recurrent Unit (GRU) since it performs similarly to Long Short-Term Memory (LSTM) but is computationally cheaper and more efficient, which is crucial to real-life HRI. The GRU network consists of a forward GRU that reads the input from left to right and a backward GRU that reads the input reversely to better model the sequential structure of the feature sequences.

\begin{align}
    \overrightarrow{h}_t &= \mathit{GRU}(\overrightarrow{x_t}, \overrightarrow{h}_{t-1}) \\
    \overleftarrow{h}_t &= \mathit{GRU}(\overleftarrow{x_t}, \overleftarrow{h}_{t-1}) \\
    h_t &= \overrightarrow{h}_t \oplus \overleftarrow{h}_t
\end{align}
where $x_t$ and $h_t$ are the input feature vector and the hidden state at time $t$ respectively, and $\oplus$ is a concatenation operation. 

Since the final result is not determined until a dialogue (six turn pairs) is over, the ensemble learning has two stages. In the first stage, there are six probability distributions in accordance to six turn-pairs from the three utterance-level classifiers. An averaging operation is conducted on the six probability distributions of each classifier, respectively, to obtain three AD detection results. Meanwhile, the interactivity-based classifier generates one AD detection result. In the second stage, a majority vote is applied on the four results, to produce the final diagnosis output.

\section{Discussion}
In this paper, we propose a novel diagnosis architecture consisting of an ensemble AD detection module and a proactive listener module. The ensemble AD detection module integrates four classifiers which are based on audio, language, disfluency, and interactivity, respectively, utilizing both utterance and dialogue information for diagnosis. This module resolves the problem of variable spontaneous dialogue by using a two-stage ensemble learning approach to hierarchically balance the effects of four classifiers. The proactive listener module categorizes user speech into three classes: question, statement, and silence, for which particular response types are generated. This approach overcomes the limitation of small-amount speech samples and breakdown dialogues. With the proposed architecture, conversational robots are expected to be applied in healthcare as a solution to the shortage of medical professionals in the aging society.

A limitation we need to take into consideration is that models trained on ``normal'' speech don't necessarily extend to disordered speech. Each component will still need to be evaluated on AD speech, especially the ones which rely on ASR, such as disfluency and dialogue act. It might be necessary to rely more on prosody using complex models \cite{b46}. We need to keep in mind that how good the state-of-the-art of each of the components is. In our future work, we plan to firstly use the dementia dialogue database ``Carolina Conversation Collection'' \cite{b47} to analyze audio, language, disfluency, and interactivity and model the classifiers. An experimental evaluation using the proactive robotic listener will be conducted as our long-term goal.


\end{document}